\RequirePackage[hyphens]{url}
\documentclass[11pt,a4paper]{article}
\usepackage[hyperref]{naaclhlt2019}
\usepackage{times}
\usepackage{latexsym}

\usepackage{url}
\usepackage{haldefs}
\usepackage{booktabs}
\usepackage{graphicx}  
\usepackage{subfigure}
\usepackage{xspace}
\usepackage{xcolor}
\usepackage[inline]{enumitem}
\usepackage{mathtools}
\usepackage{graphicx}
\usepackage{amsfonts}

\aclfinalcopy 
\def\aclpaperid{886} 

\def\equationautorefname~#1\null{Eq~#1\null}
\renewcommand{\sectionautorefname}{\S\kern-0.2em}
\renewcommand{\subsectionautorefname}{\S\kern-0.2em}
\renewcommand{\subsubsectionautorefname}{\S\kern-0.2em}

\newcommand{\mixer}{\textsc{Mixer}\xspace}
\newcommand{\utility}{\textsc{Utility}\xspace}
\newcommand{\bleu}{\textsc{Bleu}\xspace}
\newcommand{\rouge}{\textsc{Rouge}\xspace}
\newcommand{\meteor}{\textsc{Meteor}\xspace}
\newcommand{\diversity}{\textsc{Diversity}\xspace}
\newcommand{\reinforce}{\textsc{Reinforce}\xspace}

\renewcommand\cite\citep 

\title{Answer-based Adversarial Training for\\Generating Clarification Questions}

\author{Sudha Rao\Thanks{This research performed when the author was still at University of Maryland, College Park.} \\
Microsoft Research, Redmond \\
{\tt Sudha.Rao@microsoft.com} \\ \And
Hal Daum\'e III \\
University of Maryland, College Park \\
Microsoft Research, New York City\\
{\tt me@hal3.name} }

\date{}

\begin{document}
\maketitle
\begin{abstract}
We present an approach for generating clarification questions with the goal of eliciting new information that would make the given textual context more complete. 
We propose that modeling hypothetical answers (to clarification questions) as latent variables can guide our approach into generating more useful clarification questions.
We develop a Generative Adversarial Network (GAN) where the generator is a sequence-to-sequence model and the discriminator is a utility function that models the value of updating the context with the answer to the clarification question. We evaluate on two datasets, using both automatic metrics and human judgments of usefulness, specificity and relevance, showing that our approach outperforms both a retrieval-based model and ablations that exclude the utility model and the adversarial training.
\end{abstract}

\section{Introduction} \label{sec:intro}

A goal of natural language processing is to develop techniques that enable machines to process naturally occurring language.
However, not all language is clear and, as humans, we may not always understand each other \citep{grice1975logic};
in cases of gaps or mismatches in knowledge, we tend to ask questions \citep{graesser2008question}.
In this work, we focus on the task of automatically generating clarification questions: questions that ask for information that is \emph{missing} from a given linguistic context.
Our clarification question generation model builds on the sequence-to-sequence approach that has proven effective for several language generation tasks \citep{sutskever2014sequence,serban2016building,yin2016neural,du2017learning}.
Unfortunately, training a sequence-to-sequence model directly on (context, question) pairs yields questions that are highly generic\footnote{For instance, under home appliances, frequently asking ``Is it made in China?'' or ``What are the dimensions?''}, corroborating a common finding in dialog systems \citep{li2016deep}.
Our goal is to be able to generate clarification questions that are useful \emph{and} specific.

To achieve this, we begin with a recent observation of \citet{rao2018learning}, who consider the task of question reranking: a good clarification question is the one whose answer has a high \emph{utility}, which they define as the likelihood that this question would lead to an answer that will make the context more complete (\autoref{sec:utility}).
Inspired by this, we construct a model that first generates a question given a context, and then generates a hypothetical answer to that question.
Given this (context, question, answer) triple, we train a utility calculator to estimate the usefulness of this question.
We then show that this utility calculator can be generalized using ideas for generative adversarial networks \cite{goodfellow2014generative} for text \cite{yu2017seqgan}, wherein the utility calculator plays the role of the ``discriminator'' and the question generator is the ``generator'' (\autoref{sec:mixer}), which we train using the \mixer algorithm~\cite{ranzato2015sequence}. We evaluate our approach on two datasets: Amazon product descriptions (\autoref{fig:amazon_cqa}) and Stack Exchange posts (\autoref{fig:stackexchange_cqa}). Our two main contributions are:
\begin{enumerate}[noitemsep,nolistsep]
\item An adversarial training approach for generating clarification questions that models the utility of updating a context with an answer to the clarification question.
\footnote{\small Code and data: \url{https://github.com/raosudha89/clarification_question_generation_pytorch}}
\item An empirical evaluation using both automatic metrics and human judgments to show that our adversarially trained model generates questions that are more \emph{useful} and \emph{specific to the context} than all the baseline models.
\end{enumerate}
%

\begin{figure}
\centering
\small
\begin{tabular}{l | l}
\toprule
Product  & T-fal Nonstick Cookware Set,  \\
title & 18 pieces, Red \\
\midrule
Product & Easy non-stick 18pc set includes every \\
description & piece for your everyday meals. \\
 & Exceptionally durable dishwasher \\
& safe cookware for easy clean up. \\
& Durable non-stick interior. \\
& Oven safe up to 350.F/177.C \\
\midrule
 Question & Are they induction compatible? \\
 \midrule
 Answer & They are aluminium so the answer is NO. \\
\bottomrule
\end{tabular}
\caption{Sample product description from Amazon paired with a clarification question and answer.}
\label{fig:amazon_cqa}
\end{figure}



\section{Training a Clarification Question Generator} \label{sec:model} 
Our goal is to build a model that, given a context, can generate an appropriate clarification question.
Our dataset consists of (\emph{context}, \emph{question}, \emph{answer}) triples where the \emph{context} is an initial textual context, \emph{question} is the clarification question that asks about some missing information in the context and \emph{answer} is the answer to the clarification question (details in \autoref{sec:datasets}). 
Representationally, our question generator is a standard sequence-to-sequence model with attention (\autoref{sec:seq2seq}).
The learning problem is: how to train the sequence-to-sequence model to generate good clarification questions.

\begin{figure*}[!t]
\centering
 \includegraphics[scale=0.23]{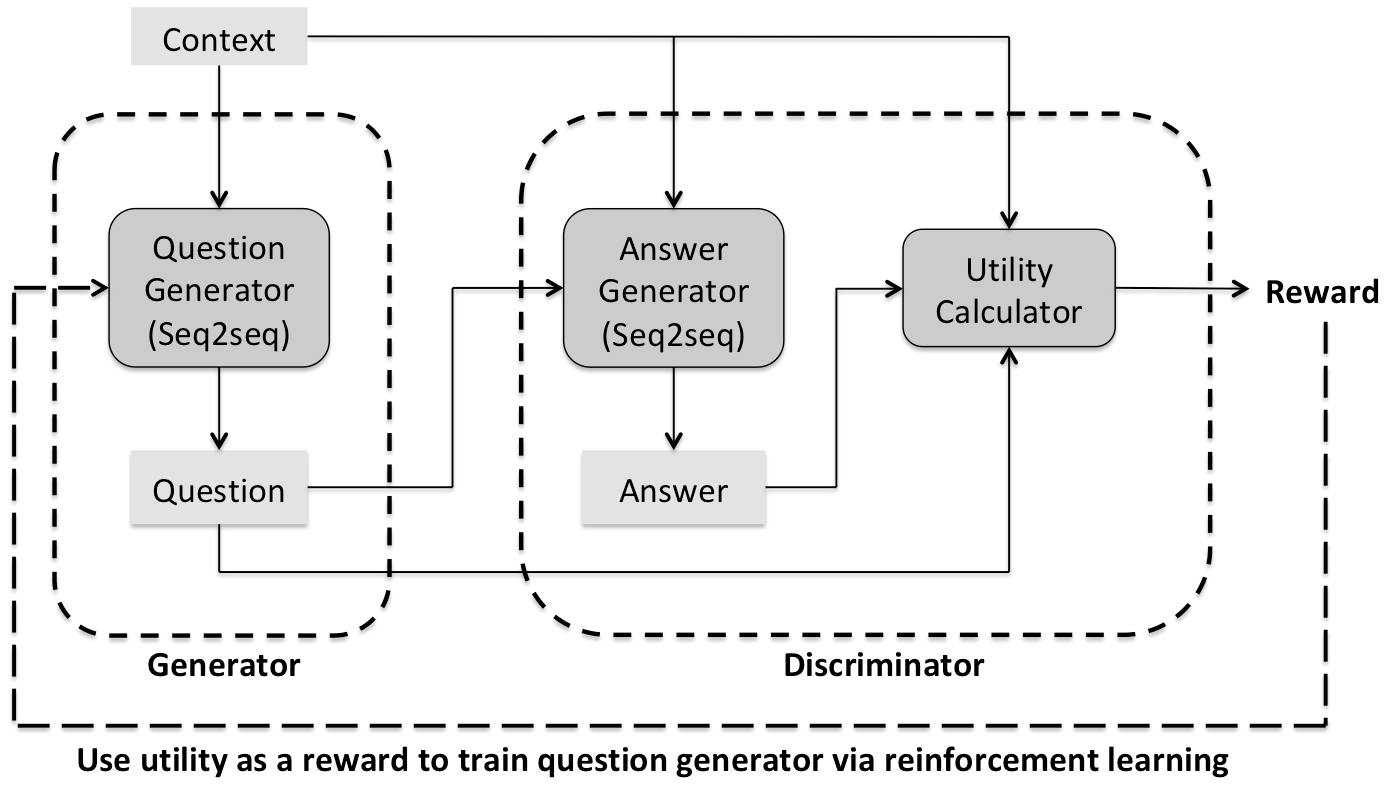}
 \caption{Overview of our GAN-based clarification question generation model (refer preamble of~\autoref{sec:model})}\label{fig:model}
\end{figure*}

An overview of our training setup is shown in \autoref{fig:model}.
Given a context, our question generator, which is a sequence-to-sequence model, outputs a question.
In order to evaluate the usefulness of this question, we then have a second sequence-to-sequence model called the ``answer generator'' that generates a hypothetical answer based on the context and the question (\autoref{sec:pretraining}).
This (context, generated question and generated answer) triple is fed into a \utility calculator, whose initial goal is to estimate the probability that this (question, answer) pair is useful in this context (\autoref{sec:utility}).
This \utility is treated as a reward, which is used to update the question generator using the \mixer \cite{ranzato2015sequence} algorithm (\autoref{sec:mixer}).
Finally, we reinterpret the answer-generator-plus-utility-calculator component as a \emph{discriminator} for differentiating between (context, true question, generated answer) triples and (context, generated question, generated answer) triples , and optimize the generator for this adversarial objective using \mixer (\autoref{sec:gan}).

\begin{figure}
\centering
\small
\begin{tabular}{l | l}
\toprule
Title & Wifi keeps dropping on 5Ghz network  \\
\midrule
Post & Recently my wireless has been very iffy at my\\
& university. I notice that I am connected to a 5Ghz \\
& network, while I am usually connected to a 2.4Ghz \\
&  everywhere else (where everything works just fine). \\
&  Sometimes it reconnects, but often I have to run \\
&  `sudo service network-manager restart`. \\
& Is it possible a kernel update has caused this?\\
\midrule
 Question & what is the make of your wifi card ? \\
 \midrule
 Answer & intel corporation wireless 7260 ( rev 73 ) \\
\bottomrule
\end{tabular}
\caption{Sample post from stackexchange.com paired with a clarification question and answer.}\label{fig:stackexchange_cqa}
\end{figure}

\subsection{Sequence-to-sequence Model for Question Generation}\label{sec:seq2seq}
We use a standard attention based sequence-to-sequence model \cite{luong2015effective} for our question generator. Given an input sequence (context) $c=(c_1,c_2, ..., c_N)$, this model generates an output sequence (question) $q =(q_1, q_2, ..., q_T)$. The architecture of this model is an encoder-decoder with attention. The encoder is a recurrent neural network (RNN) operating over the input word embeddings to compute a source context representation $\tilde{c}$. The decoder uses this source representation to generate the target sequence one word at a time:
\begin{equation}
\begin{split}
  p(q | \tilde{c}) & = \prod_{t=1}^{T} p(q_t | q_1, q_2, ..., q_{t-1}, \tilde{c_t}) \\
  & = \prod_{t=1}^T \textit{softmax}(W_s{\tilde h}_t) \quad;\quad \\
  \textrm{where } {\tilde h}_t & = \text{tanh}(W_c[\tilde{c_t}; h_t])
\end{split}
\label{eq:output-prob}
\end{equation}
In \autoref{eq:output-prob}, ${\tilde h}_t$ is the attentional hidden state of the RNN at time $t$ and $W_s$ and $W_c$ are parameters of the model.\footnote{Details are in \autoref{sec:appendix-seq2seq}.}
The predicted token $q_t$ is the token in the vocabulary that is assigned the highest probability using the softmax function. 
The standard training objective for sequence-to-sequence model is to maximize the log-likelihood of all $(c, q)$ pairs in the training data $D$ which is equivalent to minimizing the following loss, 
\begin{equation}
L_{\text{mle}}(D) =  - \sum_{(c,q) \in D} \sum_{t=1}^{T} \log p(q_t | q_1, ..., q_{t-1}, \tilde{c_t})
\end{equation}

\begin{toappendix}

\section{Sequence-to-sequence model details}\label{sec:appendix-seq2seq}

In this section, we describe some of the details of the attention based sequence-to-sequence model introduced in Section 2.1 of the main paper. 
In equation 1, ${\tilde h}_t$ is the attentional hidden state of the RNN at time $t$ obtained by concatenating the target hidden state $h_t$ and the source-side context vector $\tilde c_t$,
 and $W_s$ is a linear transformation that maps $h_t$ to an output vocabulary-sized vector.
Each attentional hidden state ${\tilde h}_t$ depends on a distinct input context vector $\tilde c_t$  computed using a global attention mechanism over the input hidden states as: 
\begin{align}
  \tilde c_t &= \sum_{n=1}^{N} a_{nt} h_n  \\
  a_{nt} & = \text{align}(h_n, h_t) \\
  & = {\exp\Big[ h_t^T W_a h_n \Big]} \Big/ {\sum_{n'}\exp\Big[ h_t^T W_a h_{n'} \Big] }
\end{align}
The attention weights $a_{nt}$ is calculated based on the alignment  score between the source hidden state $h_n$ and the current target hidden state $h_t$.

\end{toappendix}

\subsection{Training the Generator to Optimize \utility}\label{sec:mixer}

Training sequence-to-sequence models for the task of clarification question generation (with context as input and question as output) 
using maximum likelihood objective unfortunately leads to the generation of highly generic questions, such as \textit{``What are the dimensions?''} when asking questions about home appliances. 
Recently, \citet{rao2018learning} observed that the usefulness of a question can be better measured as the \emph{utility} that would be obtained if the context were updated with the answer to the proposed question. 
Following this observation, we first use a pretrained answer generator (\autoref{sec:pretraining}) to generate an answer given a context and a question. 
We then use a pretrained \utility calculator (\autoref{sec:utility} ) to predict the likelihood that the generated answer would increase the utility of the context by adding useful information to it.
Finally, we train our question generator to optimize this \utility based reward. 

Similar to optimizing metrics like \bleu and \rouge, this \utility calculator also operates on discrete text outputs, which makes optimization difficult due to non-differentiability. 
A successful recent approach dealing with the non-differentiability while also retaining some advantages of maximum likelihood training is the Mixed Incremental Cross-Entropy Reinforce \citep{ranzato2015sequence} algorithm (\mixer).
In \mixer, the overall loss $L$ is differentiated as in \reinforce \citep{williams1992simple}:
\begin{equation}
\begin{split}
L(\theta) & = - \mathbb{E}_{q^s \sim p_{\theta}}  r(q^s) \quad;\quad \\
\nabla_{\theta} L(\theta) & = - \mathbb{E}_{q^s \sim p_{\theta}} r(q^s) \nabla_{\theta} \log p_{\theta} (q^s)
\end{split}
 \label{eq:r}
\end{equation}
where $q^s$ is a random output sample according to the model $p_\theta$ and $\theta$ are the parameters of the network.
The expected gradient is then approximated using a single sample $q^s = (q^s_1, q^s_2, ..., q^s_T)$ from the model distribution ($p_{\theta}$).
In \reinforce, the policy is initialized randomly, which can cause long convergence times.
To solve this, \mixer starts by optimizing maximum likelihood for the initial $\Delta$ time steps, and slowly shifts to optimizing the expected reward from \autoref{eq:r} for the remaining $(T - \Delta)$ time steps.


In our model, for the initial $\Delta$ time steps, we minimize $L_{\text{mle}}$ and for the remaining steps, we minimize the following \utility-based loss: 
\begin{equation}
\small
\begin{split}
L_{\text{max-utility}} =  - (r(q^p) - r(q^b)) \sum_{t=1}^{T} \log p(q_t | q_1, ..., q_{t-1}, \tilde{c_t})
\end{split}
\end{equation}
where $r(q^p)$ is the \utility based reward on the predicted question and $r(q^b)$ is a baseline reward introduced to reduce the high variance otherwise observed when using \reinforce.
To estimate this baseline reward, we take the idea from the self-critical training approach \citet{rennie2017self} where the baseline is estimated using the reward obtained by the current model under greedy decoding during test time. We find that this approach for baseline estimation stabilizes our model better than the approach used in \mixer. 



\subsection{Estimating \utility from Data}\label{sec:utility}

Given a (context, question, answer) triple, \citet{rao2018learning} introduce a utility calculator $\utility(c, q, a)$ to calculate the value of updating a context $c$ with the answer $a$ to a clarification question $q$.
They use the utility calculator to estimate the probability that an \emph{answer} would be a meaningful addition to a context.
They treat this as a binary classification problem where the positive instances are the true (context, question, answer) triples in the dataset whereas the negative instances are contexts paired with a random (question, answer) from the dataset. 
Following  \citet{rao2018learning}, we model our \utility calculator by first embedding the words in $c$ and then using an LSTM (long-short term memory) \citep{hochreiter1997long} to generate a neural representation $\bar{c}$ of the context by averaging the output of each of the hidden states. 
Similarly, we obtain neural representations $\bar{q}$ and $\bar{a}$ of $q$ and $a$ respectively using a question and an answer LSTM models. 
Finally, we use a feed forward neural network $F_{\utility} (\bar{c}, \bar{q}, \bar{a})$ to predict the usefulness of the question.

\subsection{\utility GAN for Clarification Question Generation}\label{sec:gan}

The \utility calculator trained on true vs random samples from real data (as described in the previous section) can be a weak reward signal for questions generated by a model due to the large discrepancy between the true data and the model's outputs.
In order to strengthen the reward signal, we reinterpret the \utility calculator (coupled with the answer generator) as a discriminator in an adversarial learning setting.
That is, instead of taking the \utility calculator to be a fixed model that outputs the expected quality of a (question, answer) pair, we additionally optimize it to distinguish between true (question, answer) pairs and model-generated ones.
This reinterpretation turns our model into a form of a generative adversarial network (GAN) \citep{goodfellow2014generative}.

GAN is a training procedure for ``generative'' models that can be interpreted as a game between a generator and a discriminator.
The generator is a model $g \in \mathcal{G}$ that produces outputs (in our case, questions).
The discriminator is another model $d \in \mathcal{D}$ that attempts to classify between true outputs and model-generated outputs.
The goal of the generator is to generate data such that it can fool the discriminator; the goal of the discriminator is to be able to successfully distinguish between real and generated data. In the process of trying to fool the discriminator, the generator produces data that is as close as possible to the real data distribution.
Generically, the GAN objective is:
\begin{equation}
\small
\begin{split}
  L_{\text{GAN}}(\mathcal{D}, \mathcal{G}) =
  \max_{d \in \mathcal{D}} \min_{g \in \mathcal{G}} & \mathbb{E}_{x \sim \hat p} \log d(x) + \\
                                            &  \mathbb{E}_{z \sim p_z} \log(1 - d(g(z)))
\end{split}                                      
\end{equation}
where $x$ is sampled from the true data distribution $\hat p$, and $z$ is sampled from a prior defined on input noise variables $p_{z}$.

Although GANs have been successfully used for image tasks, training GANs for text generation is challenging due to the discrete nature of outputs in text. The discrete outputs from the generator make it difficult to pass the gradient update from the discriminator to the generator. Recently, \citet{yu2017seqgan} proposed a sequence GAN model for text generation to overcome this issue. They treat their generator as an agent and use the discriminator as a reward function to update the generative model using reinforcement learning techniques. 
Our GAN-based approach is inspired by this sequence GAN model with two main modifications: a) We use \mixer algorithm as our generator (\autoref{sec:mixer}) instead of a purely policy gradient approach; and b) We use \utility calculator (\autoref{sec:utility}) as our discriminator instead of a convolutional neural network (CNN).

Theoretically, the discriminator should be trained using (context, true question, true answer) triples as positive instances and (context, generated question, generated answer) triples as the negative instances.
However, we find that training a discriminator using such positive instances makes it very strong since the generator would have to not only generate real looking questions but also generate real looking answers to fool the discriminator. 
Since our main goal is question generation and since we use answers only as latent variables, we instead use (context, true question, \emph{generated answer}) as our positive instances where we use the pretrained answer generator to get the \emph{generated answer} for the true question.
Formally, our objective function is:
\begin{equation}
\footnotesize
\begin{split}
  L_{\text{GAN-U}}(\mathcal{U}, \mathcal{M}) = &
  \max_{u \in \mathcal{U}} \min_{m \in \mathcal{M}} \mathbb{E}_{q \sim \hat p} \log u(c, q, \mathcal{A}(c, q)) + \\
                                             & \mathbb{E}_{c \sim \hat p} \log(1 - u(c, m(c), \mathcal{A}(c, m(c))))
\end{split}                                      
\end{equation}
\normalsize
where $\mathcal{U}$ is the \utility discriminator, $\mathcal{M}$ is the \mixer generator, $\hat p$ is our data of (context, question, answer) triples and $\mathcal{A}$ is the answer generator.

\subsection{Pretraining}\label{sec:pretraining}

\textbf{Question Generator.} We pretrain our question generator using the sequence-to-sequence model (\autoref{sec:seq2seq}) to maximize the log-likelihood of all (context, question) pairs in the training data. 
Parameters of this model are updated during adversarial training. 

\textbf{Answer Generator.} We pretrain our answer generator using the sequence-to-sequence model (\autoref{sec:seq2seq}) to maximize the log-likelihood of all ([context+question], answer) pairs in the training data. 
Parameters of this model are kept fixed during the adversarial training.\footnote{We leave the experimentation of updating parameters of answer generator during adversarial training to future work.} 

\textbf{Discriminator.} In our \utility GAN model (\autoref{sec:gan}), the discriminator is trained to differentiate between true and generated questions. 
However, since we want to guide our \utility based discriminator to also differentiate between true (``good'') and random (``bad'') questions, 
we pretrain our discriminator in the same way we trained our \utility calculator.
For positive instances, we use a context and its true question, answer from the training data and for negative instances, we use the same context but randomly sample a question from the training data (and use the answer paired with that random question).


\section{Experimental Results} \label{sec:experiments} 
We base our experimental design on the following research questions:
\begin{enumerate}[noitemsep,nolistsep]
\item Do generation models outperform simpler retrieval baselines?
\item Does optimizing the \utility reward improve over maximum likelihood training?
\item Does using adversarial training improve over optimizing the pretrained \utility?
\item How do the models perform when evaluated for nuances such as specificity \& usefulness?
\end{enumerate}

\subsection{Datasets} \label{sec:datasets}

We evaluate our model on two datasets. 

\textbf{Amazon.}
In this dataset, \emph{context} is a product description on amazon.com combined with the product title, \emph{question} is a clarification question asked to the product and \emph{answer} is the seller's (or other users') reply to the question. 
To obtain these data triples, we combine the Amazon question-answering dataset \cite{mcauley2016addressing} with the Amazon reviews dataset  \cite{mcauley2015image}.
We show results on the \texttt{Home \& Kitchen} category of this dataset since it contains a large number of questions and is relatively easier for human-based evaluation. 
It consists of $19,119$ training, $2,435$ tune and $2,305$ test examples (product descriptions), with 3 to 10 questions (average: 7) per description.

\textbf{Stack Exchange.}
In this dataset, \emph{context} is a post on stackexchange.com combined with the title, \emph{question} is a clarification question asked in the comments section of the post and 
\emph{answer} is either the update made to the post in response to the question or the author's reply to the question in the comments section.
\citet{rao2018learning} curated a dataset of $61,681$ training, $7,710$ tune and $7,709$ test such triples from three related subdomains on stackexchage.com (askubuntu, unix and superuser).
Additionally, for 500 instances each from the tune and the test set, their dataset includes 1 to 6 other questions identified as valid questions by expert human annotators from a pool of candidate questions. 

\subsection{Baselines and Ablated Models}

We compare three variants (ablations) of our proposed approach, together with an information retrieval baseline:

\textbf{GAN-Utility} is our full model which is a \utility calculator based GAN training (\autoref{sec:gan}) including the \utility discriminator and the \mixer question generator.\footnote{Experimental details are in \autoref{sec:appendix-exp-details}.}

\textbf{Max-Utility} is our reinforcement learning baseline where the pretrained question generator model is further trained to optimize the  \utility reward (\autoref{sec:mixer}) without the adversarial training.

\textbf{MLE} is the question generator model pretrained on context, question pairs using maximum likelihood objective (\autoref{sec:seq2seq}).

\textbf{Lucene}\footnote{\url{https://lucene.apache.org/}} is our information retrieval baseline similar to the Lucene baseline described in \citet{rao2018learning}. 
Given a context in the test set, we use Lucene, which is a TF-IDF based document ranker, to retrieve top 10 contexts that are most similar to the given context in the train set. 
We randomly choose a question from the human written questions paired with these 10 contexts in the train set to construct our Lucene baseline\footnote{For the Amazon dataset, we ignore questions asked to products of the same brand as the given product since Amazon replicates questions across same brand allowing the true question to be included in that set.}. 

\begin{toappendix}

 \section{Experimental Details}\label{sec:appendix-exp-details}
 
  In this section, we describe the details of our experimental setup. 
  
 We preprocess all inputs (context, question and answers) using tokenization and lowercasing. 
 We set the max length of context to be 100, question to be 20 and answer to be 20. 
 We test with context length 150 and 200 and find that the automatic metric results are similar as that of context length 100 but the experiments take much longer. Hence, we set the max context length to be 100 for all our experiments. 
 Similarity, we find that an increased length of question and answer yields similar results with increased experimentation time. 
 
 Our sequence-to-sequence model (Section 2.1) operates on word embeddings which are pretrained on in domain data using Glove \citep{pennington2014glove}.
 As frequently used in previous work on neural network modeling, we use an embeddings of size 200 and a vocabulary with cut off frequency set to 10. 
 During train time, we use teacher forcing \citep{williams1989learning}. 
 During test time, we use beam search decoding with beam size 5.
 
 We use a hidden layer of size two for both the encoder and decoder recurrent neural network models with size of hidden unit set to 100. 
 We use a dropout of 0.5 and learning ratio of 0.0001. 
 In the \mixer model, we start with $\Delta = T$ and decrease it by 2 for every epoch (we found decreasing $\Delta$ to 0 is ineffective for our task, hence we stop at 2).

\end{toappendix}

\subsection{Evaluation Metrics} \label{sec:methodology}

We evaluate initially with automated evaluation metrics, and then more substantially with crowdsourced human judgments.

\subsubsection{Automatic Metrics}
\textbf{Diversity}, which calculates the proportion of unique trigrams in the output to measure the diversity as commonly used to evaluate dialogue generation \citep{li2016deep}.\\
\textbf{\bleu} \cite{papineni2002bleu} \footnote{\small\url{https://github.com/moses-smt/mosesdecoder/blob/master/scripts/generic/multi-bleu.perl}}, which evaluates n-gram precision between the output and the references. \\
\textbf{\meteor} \cite{banerjee2005meteor}, which is similar to \bleu but includes stemmed and synonym matches to measure similarity between the output and the references.

\subsubsection{Human Judgements}

We use Figure-Eight\footnote{\url{https://www.figure-eight.com}}, a crowdsourcing platform, to collect human judgements. 
Each judgement\footnote{We paid crowdworkers 5 cents per judgment and collected five judgments per question.} consists of showing the crowdworker a context and a generated question and asking them to evaluate the question along following axes:

\noindent \textbf{Relevance}: We ask \textit{``Is the question on topic?"}
 and let workers choose from: Yes (1) and No (0)

\noindent \textbf{Grammaticality}: We ask \textit{``Is the question grammatical?"} and let workers choose from: Yes (1) and No (0)

\noindent \textbf{Seeking new information}: We ask \textit{``Does the question ask for new information currently not included in the description?''} and let workers choose from: Yes (1) and No (0)

\noindent \textbf{Specificity}: We ask \textit{``How specific is the question?''} and let workers choose from: 
 \begin{description}[noitemsep,nolistsep]
 \item 4: Specific pretty much only to this product (or same product from different manufacturer)
 \item 3: Specific to this and other very similar products
 \item 2: Generic enough to be applicable to many other products of this type
 \item 1: Generic enough to be applicable to any product under Home and Kitchen
 \item 0: N/A (Not applicable) i.e. Question is not on topic OR  is incomprehensible
 \end{description}

\noindent \textbf{Usefulness}: We ask \textit{``How useful is the question to a potential buyer (or a current user) of the product?''} and let workers choose from:
 \begin{description}[noitemsep,nolistsep]
 \item 4: Useful enough to be included in the product description
 \item 3: Useful to a large number of potential buyers (or current users)
 \item 2: Useful to a small number of potential buyers (or current users)
 \item 1: Useful only to the person asking the question
 \item 0: N/A (Not applicable) i.e. Question is not on topic OR is incomprehensible  OR is not seeking new information
 \end{description}

\subsubsection{Inter-annotator Agreement}

\autoref{tab:agreement} shows the inter-annotator agreement (reported by Figure-Eight as confidence\footnote{\small\url{https://success.figure-eight.com/hc/en-us/articles/201855939-How-to-Calculate-a-Confidence-Score}}) on each of the above five criteria.
Agreement on \textit{Relevance}, \textit{Grammaticality} and \textit{Seeking new information} is high. This is not surprising given that these criteria are not very subjective.
On the other hand, the agreement on usefulness and specificity is quite moderate since these judgments can be very subjective.

\begin{table}[t]
\begin{tabular}{l c}
\bf Criteria & \bf Agreement \\
\midrule
Relevance & 0.92 \\
Grammaticality & 0.92 \\
Seeking new information & 0.84 \\
Usefulness & 0.65 \\
Specificity & 0.72 \\
\end{tabular}
\caption{Inter-annotator agreement on the five criteria used in human-based evaluation.}\label{tab:agreement}
\vspace{-1.1em}
\end{table}

Since the inter-annotator agreement on the usefulness criteria was particularly low, in order to reduce the subjectivity involved in the fine grained annotation,
we convert the range [0-4] to a more coarse binary range [0-1] by mapping the scores 4 and 3 to \textbf{1} and the scores 2, 1 and 0 to \textbf{0}. 

\subsection{Automatic Metric Results} \label{sec:results}

\begin{table*}[t]
\centering
\begin{tabular}{l  c c c  |  c c c}
  \toprule
& \multicolumn{3}{c|}{\texttt{Amazon}} & \multicolumn{3}{c}{\texttt{StackExchange}} \\
Model & \diversity & \bleu & \meteor & \diversity & \bleu & \meteor \\
\midrule
Reference & 0.6934 & --- & --- & 0.7509 & --- & ---  \\
Lucene & 0.6289 & 4.26 & 10.85 & 0.7453 & 1.63 & 7.96  \\
\midrule
MLE & 0.1059 & \bf 17.02 & 12.72 &  0.2183 & 3.49 & 8.49  \\
Max-Utility & 0.1214 & 16.77 & 12.69 &  \bf 0.2508 & 3.89 & 8.79  \\
GAN-Utility & \bf 0.1296 & 15.20 & \bf 12.82 &  0.2256 & \bf 4.26 & \bf 8.99 \\
\bottomrule
\end{tabular}
\caption{\diversity as measured by the proportion of unique trigrams in model outputs. Bigrams and unigrams follow similar trends. \bleu and \meteor scores using up to 10 references for the Amazon dataset and up to six references for the StackExchange dataset. Numbers in bold are the highest among the models. All results for Amazon are on the entire test set whereas for StackExchange they are on the 500 instances of the test set that have multiple references.}\label{tab:results}
\end{table*}

\autoref{tab:results} shows the results on the two datasets when evaluated according to automatic metrics. 

In the Amazon dataset, GAN-Utility outperforms all ablations on \diversity, suggesting that it produces more diverse outputs.
Lucene, on the other hand, has the highest \diversity since it consists of human written questions, which tend to be more diverse because they are much longer compared to model generated questions.
This comes at the cost of lower match with the reference as visible in the \bleu and \meteor scores. 
In terms of \bleu and \meteor, there is inconsistency.
Although GAN-Utility outperforms all baselines according to \meteor, the fully ablated MLE model has a higher \bleu score. 
This is because \bleu score looks for exact n-gram matches and since MLE produces more generic outputs, it is much more likely that it will match one of 10 references compared to the specific/diverse outputs of GAN-Utility, since one of those ten is highly likely to itself be generic.

In the StackExchange dataset GAN-Utility outperforms all ablations on both \bleu and \meteor. 
Unlike in the Amazon dataset, MLE does not outperform GAN-Utility in \bleu. This is because the MLE outputs in this dataset are not as generic as in the amazon dataset due to the highly technical nature of contexts in StackExchange.  
As in the Amazon dataset, GAN-Utility outperforms MLE on \diversity. Interestingly, the Max-Utility ablation achieves a higher \diversity score than GAN-Utility. 
On manual analysis we find that Max-Utility produces longer outputs compared to GAN-Utility but at the cost of being less grammatical.

\subsection{Human Judgements Analysis} \label{sec:analysis}

\autoref{tab:analysis-results} shows the numeric results of human-based evaluation performed on the reference and the system outputs on 300 random samples from the test set of the Amazon dataset.\footnote{We could not ask crowdworkers evaluate the StackExchange data due to its highly technical nature.}
All approaches produce relevant and grammatical questions.
All models are all equally good at seeking new information, but are weaker than Lucene, which performs better at seeking new information but at the cost of much lower specificity and lower usefulness.

\begin{figure*}[t]
    \begin{minipage}[b]{0.48\textwidth}
    \includegraphics[width=\textwidth]{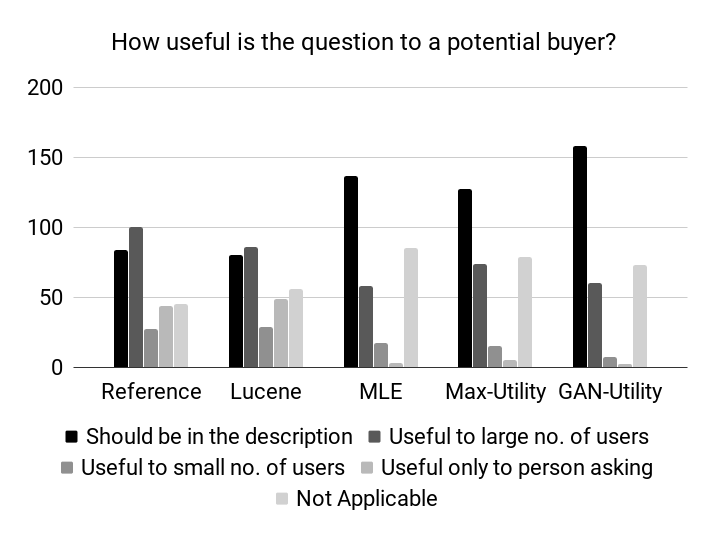}
    \caption{Human judgements on the usefulness criteria.}\label{fig:usefulness-analysis}
    \end{minipage}
	\hspace{4mm}
	\begin{minipage}[b]{0.48\textwidth}
	\includegraphics[width=\textwidth]{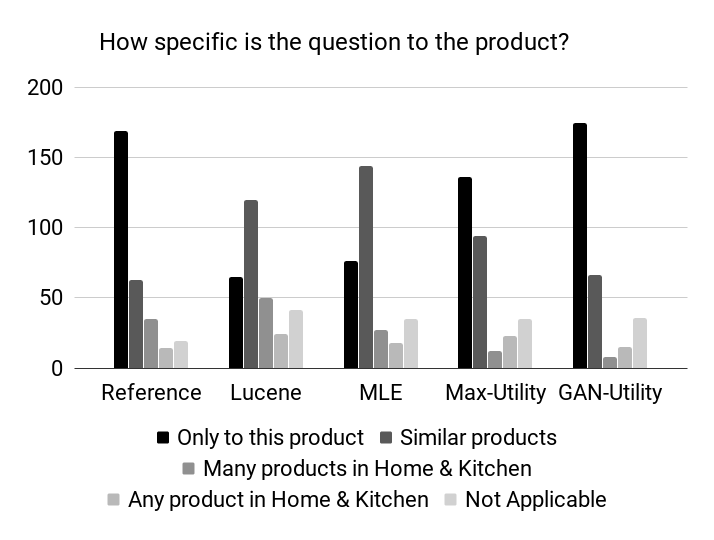}
    \caption{Human judgements on the specificity criteria.}\label{fig:specificity-analysis}
	\end{minipage}
\end{figure*}

Our full model, GAN-Utility, performs significantly better at the usefulness criteria showing that the adversarial training approach generates more useful questions. 
Interestingly, all our models produce questions that are more useful than Lucene and Reference, largely because Lucene and Reference tend to ask questions that are more often useful only to the person asking the question, making them less useful for potential other buyers (see \autoref{fig:usefulness-analysis}). 
GAN-Utility also performs significantly better at generating questions that are more specific to the product (see details in \autoref{fig:specificity-analysis}), which aligns with the higher \diversity score obtained by GAN-Utility under automatic metric evaluation. 

\autoref{tab:example-outputs} contains example outputs from different models along with their usefulness and specificity scores. 
MLE generates questions such as \textit{``is it waterproof?''} and \textit{``what is the wattage?''}, which are applicable to many other products.
Whereas our GAN-Utility model generates more specific question such as \textit{``is this shower curtain mildew resistant?''}. 
\autoref{sec:appendix-analysis} includes further analysis of system outputs on both Amazon and Stack Exchange datasets.

\begin{table*}[t]
\centering
\begin{tabular}{lccccc}
\toprule
Model & Relevant {\tiny [0-1]} & Grammatical {\tiny [0-1]} & New Info {\tiny [0-1]} & Useful {\tiny [0-1]} & Specific {\tiny [0-4]}   \\
\midrule
Reference &  0.96 & 0.99 & 0.93 & 0.72  & 3.38 \\
\midrule
Lucene & \bf 0.90 & \bf 0.99 & \bf 0.95 & 0.68 & 2.87 \\
MLE & \bf 0.92 & \bf 0.96  & 0.85 & 0.91 & 3.05  \\
Max-Utility & \bf 0.93 & \bf 0.96 & 0.88 &  0.91 & 3.29  \\
GAN-Utility & \bf 0.94 & \bf 0.96  & 0.87 & \bf 0.96 & \bf 3.52 \\
\bottomrule
\end{tabular}
\caption{Results of human judgments on model generated questions on 300 sample Home \& Kitchen product descriptions. Numeric range corresponds to the options described in \autoref{sec:methodology}.
The difference between the bold and the non-bold numbers is statistically significant with p \textless 0.05. Reference is excluded in the significance calculation.}\label{tab:analysis-results}
\end{table*}

\begin{table*}[t]
\centering
\small
\begin{tabular}{l |  l | c c}
\toprule
Title & Raining Cats and Dogs Vinyl Bathroom \textbf{Shower Curtain} & &  \\
\midrule
Product & This adorable shower curtain measures 70 by 72  & & \\
Description & inches and is sure to make a great gift!  & & \\
\midrule
& & Usefulness {\tiny [0-4]} & Specificity {\tiny [0-4]}  \\
Reference & does the vinyl smells? & 3 & 4 \\
Lucene & other than home sweet home , what other sayings on the shower curtain ? & 2 & 4 \\
MLE & is it waterproof ? & 4 & 2 \\
Max-Utility  & is this shower curtain mildew ? & 0 & 0 \\
GAN-Utility & is this shower curtain mildew resistant ?  & 4 & 4 \\
\bottomrule
& & & \\

Title & PURSONIC HF200 Pedestal \textbf{Bladeless Fan \& Humidifier} All-in-one & &  \\
\midrule
Product & The first bladeless fan to incoporate a humidifier! ,  & & \\
Description & This product operates solely as a fan, a humidifier or both simultaneously. & & \\
& Atomizing function via ultrasonic. 5.5L tank lasts up to 12 hours. & & \\ 
\midrule
& & Usefulness {\tiny [0-4]} & Specificity {\tiny [0-4]}  \\
Reference & i can not get the humidifier to work & 1 & 2\\
Lucene & does it come with the vent kit  & 3 & 3 \\
MLE & what is the wattage of this fan ? & 4 & 2\\
Max-Utility & is this battery operated ? & 3 & 2 \\
GAN-Utility & does this fan have an automatic shut off ? & 4 & 4 \\
\bottomrule

\end{tabular}
\caption{Example outputs from each of the systems for two product descriptions along with the usefulness and the specificity score given by human annotators.}\label{tab:example-outputs}
\end{table*}

\begin{toappendix}

\section{Analysis of System Outputs}\label{sec:appendix-analysis}

\subsection{Amazon Dataset}

\autoref{tab:amazon-example-outputs} shows the system generated questions for three product descriptions in the Amazon dataset.

In the first example, the product is a shower curtain. 
The Reference question is specific and highly useful.
Lucene, on the other hand, picks a moderately specific (``how to clean it?'') but useful question. 
MLE model generates a generic but useful ``is it waterproof?''.
Max-Utility generates comparatively a much longer question but in doing so loses out on relevance. 
This behavior of generating two unrelated sentences is observed quite a few times in both Max-Utility and GAN-Utility models. 
This suggests that these models, in trying to be very specific, end up losing out on relevance.
In the same example, GAN-Utility also generates a fairly long question which, although awkwardly phrase, is quite specific and useful. 

In the second example, the product is a Duvet Cover Set. 
Both Reference and Lucene questions here are examples of questions that are pretty much useful only to the person asking the question. 
We find many such questions in both Reference and Lucene outputs which is the main reason for the comparatively lower usefulness scores for their outputs. 
All three of our models generate irrelevant questions since the product description explicitly says that the set is full size. 

In the last example, the product is a set of mopping clothes.
Reference question is quite specific but has low usefulness.
Lucene picks an irrelevant question.
MLE and Max-Utility generate highly specific and useful questions.
GAN-Utility generates an ungrammatical question by repeating the last word many times. 
We observe this behavior quite a few times in the outputs of both Max-Utility and GAN-Utility models suggesting that our sequence-to-sequence models are not very good at maintaining long range dependencies.

\begin{table*}[t]
\centering
\small
\begin{tabular}{l l c c}
\toprule

Title & Extra Wide \textbf{Shower Curtain} for a Clawfoot Tub/opaque with Magnets & &  \\
\midrule
Product  & Approximately 180" Wide 70" Long.  & & \\
& Extra Wide to keep the water and heat in. & & \\
Description & Opaque in color to let the light in. & & \\
& Can be used as a liner or the curtain itself. & & \\
\midrule
& & Usefulness {\tiny [0-4]} & Specificity {\tiny [0-4]}\\
Reference & does this curtain come with hooks? & 4 & 4\\
Lucene & how to clean it ? i had it over 10 months now . & 3 & 3 \\
& and some mildew grow on it  & & \\
MLE  & is it waterproof ? & 1 & 2 \\
Max-Utility  & what is the diameter of the bottom of the shower ?   & 0 & 0 \\
 & i have a kohler shower curtain . & & \\
GAN-Utility & does this curtain have suction feet on the bottom of the shower curtain ? & 3 & 4  \\
\bottomrule
& & & \\

Title & Chezmoi Collection 7-piece Chic Ruched White \textbf{Duvet Cover Set}, & &  \\
& Full Size (with Throw Pillows)   & & \\
\midrule
Product & Create a world of romance with the elegantly-styled, & & \\
Description & ruching,and luxurious all white duvet cover set.   & & \\
\midrule
& & Usefulness {\tiny [0-4]} & Specificity {\tiny [0-4]}  \\
Reference & How long will it take to ship this to my pin code? & 1 & 1\\
Lucene & and can you use the duvet as is ?   & 1 &  4\\
& if not what shall i purchase to put in it for winter or fall ? & & \\
MLE & what are the dimensions of the king size ? & 0 & 0\\
Max-Utility  & what are the dimensions of the king size ? & 0 & 0 \\
GAN-Utility & does the king size come with a duvet cover & 0 & 0 \\
& or do you have to buy a king duvet ?  & & \\
\bottomrule
& & & \\
Title & Microfiber 3-Pack, \textbf{Pro-Clean Mopping Cloths} for & &  \\
& Braava Floor Mopping Robot  & & \\
\midrule
Product &  Braavas textured Pro-Clean microfiber mopping cloths remove dirt and& & \\
Description &  hair from your floors. The cloths can be washed and used hundreds   & & \\
& of times. They are compatible with all Braava models,  including the & & \\
& Pro-Clean Reservoir Pad. Each cloth is easy to attach and remove from  & & \\
&  the magnetic cleaning pad. & & \\
\midrule
& & Usefulness {\tiny [0-4]} & Specificity {\tiny [0-4]}  \\
Reference & do i have to use a new cloth every time i want to clean my floor? & 2 & 4\\
&  \$5/\$6 seems expensive per clean & & \\
Lucene & do they remove pet odor ?  & 0  & 0 \\
MLE & will these work with the scooba ? & 3 & 3\\
Max-Utility  & do these cloths work on hardwood floors ? & 3 & 4 \\
GAN-Utility & will this work with the scooba mop mop mop mop mop mop mop & 0 & 0 \\
\bottomrule


\end{tabular}
\caption{Example outputs from each of the systems for three product descriptions from the Home \& Kitchen category of the Amazon dataset. }\label{tab:amazon-example-outputs}
\end{table*}

\subsection{Stack Exchange Dataset}

\autoref{tab:se-example-outputs} includes system outputs for three posts from the Stack Exchange dataset.

The first example is of a post where someone describes their issue of not being able to recover from their boot. 
Reference and Lucene questions are useful.
MLE generates a generic question that is not very useful.
Max-Utility generates a useful question but has slight ungrammaticality in it.
GAN-Utility, on the other hand, generates a specific and an useful question.

In the second example, again Reference and Lucene questions are useful.
MLE generates a generic question.
Max-Utility and GAN-Utility both generate fairly specific question but contain unknown tokens. 
The Stack Exchange dataset contains several technical terms leading to a long tail in the vocabulary.
Owing to this, we find that both Max-Utility and GAN-Utility models generate many instances of questions with unknown tokens.

In the third example, the Reference question is very generic. Lucene asks a relevant question.
MLE again generates a generic question.
Both Max-Utility and GAN-Utility generate specific and relevant questions. 

\begin{table*}[t]
\centering
\small
\begin{tabular}{l l }
\toprule
Title & how can i recover my boot ?\\
\midrule
Post & since last week i am trying to recover my boot . \\
& after the last update for ubuntu 12.04 i lost it and nobody could help me , \\
& i used boot-repair but there are problems with dependences , which ca n't be fix .\\
& i will be very grateful if somebody could help me . \\
\midrule
Reference &  what happens when you try to boot ? \\
Lucene & can you get into bios ?   \\
MLE & how much ram do you have ?   \\
Max-Utility  & do you have a swap partition partition ? \\
GAN-Utility & what happens when you try to boot into safe mode ?  \\
\bottomrule
& \\
Title & packages have unmet dependencies when trying to install anything \\
\midrule
Post & i 'm running ubuntu 14.04 lts . ive recently run into this problem with several applications , \\
& it seems to happen to anything i need to update or install . i just gave up the first 2 or 3 times this \\
& happened after trying a few solutions to no avail , but now i 'm having the same issue with steam \\
& trying to update , which i use quite a lot . ive looked through dozens of posts about similar issues \\
& and tried a lot of solutions and nothing seems to work.  \\
Reference & sudo dpkg -reconfigure all ? ?\\
Lucene & if you use the graphical package manager , does n't add the required packages automatically ?\\
MLE & how long did you wait ? \\
Max-Utility & can you post the output of `apt-cache policy UNK ?\\
GAN-Utility & can you post a screenshot of the output of `sudo apt-get install UNK \\
\bottomrule
& \\
Title & full lubuntu installation on usb ( uefi capable )  \\
\midrule
Post &  i want to do a full lubuntu installation on a usb stick that can be booted in uefi mode. \\
& i do not want persistent live usb but a full lubuntu installation ( which happens to live on a usb stick ) \\
& and that can boot fromanyuefi-capable computer ...\\
\midrule
Reference & hello and welcome on askubuntu . could you please clarify what you want ?\\
Lucene & so , ubuntu was installed to the pen drive ? \\
MLE & which version of ubuntu ? \\
Max-Utility & do you have a live cd or usb stick ? \\
GAN-Utility & what is the model of the usb stick ? \\
\bottomrule
\end{tabular}
\caption{Example outputs from each of the systems for three posts of the Stack Exchange dataset. }\label{tab:se-example-outputs}
\end{table*}
\end{toappendix}


\section{Related Work} \label{sec:related} 
\textbf{Question Generation.}
Most previous work on question generation has been on generating reading comprehension style questions i.e. questions that ask about information present in a given text \citep{heilman2011automatic, rus2010first, rus2011question, duan2017question}. Our goal, on the other hand, is to generate questions whose answer cannot be found in the given text. 
Outside reading comprehension questions, 
\citet{liu2010automatic} use templated questions to help authors write better related work sections whereas we generate questions to fill information gaps. 
\citet{labutov2015deep} use crowdsourcing to generate question templates whereas we learn from naturally occurring questions. 
\citet{mostafazadeh2016generating,mostafazadeh2017image} generate natural and engaging questions, given an image (and some initial text). 
Whereas, we generate questions specifically for identifying missing information.
\citet{stoyanchev2014towards} generate clarification questions to resolve ambiguity caused by speech recognition failures during dialog, whereas we generate clarification questions to resolve ambiguity caused by missing information. 
The recent work most relevant to our work is by \citet{rao2018learning}. They build a model which given a context and a set of candidate clarification questions, ranks them in a way that more useful clarification questions would be higher up in the ranking. In our work, we build on their ideas to propose a model that generates (instead of ranking) clarification questions given a context. \\


\textbf{Neural Models and Adversarial Training for Text Generation.}
Neural network based models have had significant success at a variety of text generation tasks, including machine translation \citep{BahdanauCB15,luong2015effective}, summarization \citep{nallapati2016abstractive}, dialog \citep{bordes2016learning,li2016persona,serban2017hierarchical}, textual style transfer \cite{jhamtani2017shakespearizing,RaoT18} and question answering \citep{yin2016neural,serban2016building}. 
Our task is most similar to dialog, in which a wide variety of possible outputs are acceptable, and where lack of specificity in generated outputs is common.
We addresses this challenge using an adversarial network approach \citep{goodfellow2014generative}, a training procedure that can generate natural-looking outputs, which have been effective for natural image generation \citep{denton2015deep}. 
Due to the challenges in optimizing over discrete output spaces like text, \citet{yu2017seqgan} introduced a Seq(uence)GAN approach where they overcome this issue by using \reinforce to optimize. 
Our GAN-Utility model is inspired by the SeqGAN model where we replace their policy gradient based generator with a \mixer model and their CNN based discriminator with our \utility calculator. 
\citet{li2017adversarial} train an adversarial model similar to SeqGAN for generating next utterance in a dialog given a context. 
However, unlike our work, their discriminator is a binary classifier trained only to distinguish between human and machine generated utterances. 


\section{Conclusion} \label{sec:conclusion} 
In this work, we describe a novel approach to the problem of clarification question generation.
We use the observation of \citet{rao2018learning} that the usefulness of a clarification question can be measured by the value of updating a context with an answer to the question. 
We use a sequence-to-sequence model to generate a question given a context and a second sequence-to-sequence model to generate an answer given the context and the question. 
Given the (context, generated question, generated answer) triple, we calculate the utility of this triple and use it as a reward to retrain the question generator using reinforcement learning based \mixer model. 
Further, to improve upon the utility calculator, we reinterpret it as a discriminator in an adversarial setting and train both the utility calculator and the \mixer model in a minimax fashion. 
We find that our adversarial training approach produces more useful and specific questions compared to both a model trained using maximum likelihood objective and a model trained using utility reward based reinforcement learning. 

There are several avenues of future work. 
Following \citet{mostafazadeh2016generating}, we could combine text input with image input in the Amazon dataset \citep{mcauley2016addressing} to generate more relevant and useful questions. 
One significant research challenge in the space of free text generation problems when the set of possible outputs is large, is that of automatic evaluation \cite{lowe2016evaluation}: in our results we saw some correlation between human judgments and automatic metrics, but not enough to trust the automatic metrics completely.
Lastly, we hope to integrate such a question generation model into a real world platform like StackExchange or Amazon to understand the real utility of such models and to unearth additional research questions.


\section*{Acknowledgments}

We thank the three anonymous reviewers for their helpful comments and suggestions. We also thank the members of the Computational Linguistics and Information Processing (CLIP) lab at University of Maryland for helpful discussions. This work was supported by NSF grant IIS-1618193. Any opinions, findings, conclusions, or recommendations expressed here are those of the authors and do not necessarily reflect the view of the sponsors.

\bibliography{clarification_question_generation}
\bibliographystyle{acl_natbib}

\newpage
\appendix

\makeappendix

\end{document}